\documentclass[letterpaper, 10 pt, conference]{ieeeconf}  
\pdfoutput=1 

\IEEEoverridecommandlockouts                              
\usepackage[latin1]{inputenc} 
\usepackage{graphicx} 
\usepackage{epsfig} 
\usepackage{ifpdf}

\ifpdf
	\usepackage{epstopdf}
\else
\fi
\DeclareGraphicsExtensions{.eps,.png,.pdf,.jpg,.mps}

\usepackage{mathptmx} 
\usepackage{times} 
\usepackage{amsmath} 
\usepackage{amssymb}  
\usepackage[english]{babel}
\usepackage{url}
\usepackage{listings}
\usepackage{color}

\definecolor{lightgray}{gray}{0.9}

\lstdefinelanguage{KRL}%
  {morekeywords={IMPORT,PUBLIC,DEFFCT,DELAY,PTP,EXIT,PRIO,WHEN,DECL,REPEAT,EXOR,LIN_REL,%
   ENDSWITCH,ENUM,CHANNEL,B_EXOR,B_OR,ONSTART,FOR,DEFAULT,STRUC,ENDFCT,RESUME,%
   ENDLOOP,OR,DEFDAT,PTP_REL,SPL,ENDDAT,SLIN,TO,B_AND,THEN,ENDWHILE,B_NOT,%
   BRAKE,NOT,EXT,TRIGGER,RETURN,UNTIL,ENDFOR,BREAK,GOTO,ANOUT,WHILE,SCIRC,%
   AND,CASE,WAIT,IF,ANIN,CIRC,SPLINE,HALT,SIGNAL,ENDIF,ELSE,CONTINUE,PATH,%
   DISTANCE,DEF,GLOBAL,END,LIN,DO,INTERRUPT,SWITCH,LOOP,CALL,C_DIS},%
   alsoletter=.,%
   alsodigit=?,%
   sensitive=f,%
   morestring=[b]",%
   morestring=[b]',%
   morecomment=[l];%
   }[keywords,comments,strings]
\lstdefinelanguage{ANTLR}{}[]
   
\newcommand{\krl}[1]{\lstinline[breaklines=false,language=KRL]{#1}}

\title{\LARGE \bf On reverse-engineering the KUKA Robot Language}

\author{Henrik Mühe, Andreas Angerer, Alwin Hoffmann, and Wolfgang Reif%
\thanks{Andreas Angerer, Alwin Hoffmann, Henrik Mühe, and Wolfgang Reif are with the 
	Institute for Software and Systems Engineering, 
	University of Augsburg, D-86135 Augsburg, Germany. 
	E-mail of corresponding author: {\tt hoffmann@informatik.uni-augsburg.de}
	}
\thanks{This work presents results of the research project \emph{SoftRobot} which is funded by the European Union and the Bavarian government within the \emph{High-Tech-Offensive Bayern}. The project is carried out together with KUKA Roboter GmbH and MRK-Systeme GmbH.}
}

\begin{document}

\maketitle
\thispagestyle{empty}
\pagestyle{empty}

\begin{abstract}
Most commercial manufacturers of industrial robots require their robots to be programmed in a proprietary language tailored to the domain -- a typical domain-specific language (DSL). However, these languages oftentimes suffer from shortcomings such as controller-specific design, limited expressiveness and a lack of extensibility. For that reason, we developed the extensible Robotics API for programming industrial robots on top of a general-purpose language. Although being a very flexible approach to programming industrial robots, a fully-fledged language can be too complex for simple tasks. Additionally, legacy support for code written in the original DSL has to be maintained. For these reasons, we present a lightweight  implementation of a typical robotic DSL, the KUKA Robot Language (KRL), on top of our Robotics API. This work deals with the challenges in reverse-engineering the language and mapping its specifics to the Robotics API. We introduce two different approaches of interpreting and executing KRL programs: tree-based and bytecode-based interpretation.

\end{abstract}

\section{Introduction} \label{sec:introduction}

For programming industrial robots, Biggs and MacDonald distinguish between two major principles: automatic and manual programming~\cite{Biggs2003}. While automatic programming often uses programming-by-demonstration, manual programming is by far the most common technique; the developer has to explicitly specify the desired behavior of the robot in a machine-readable syntax, usually using a formal text-based or graphical language.
Hence, a straightforward idea is to program the robot directly using a general-purpose programming language. 
This approach has the major disadvantage that the developer is responsible for developing code which satisfies all necessary real-time constraints of the robot controller. Any errors in the code may not only lead to a failure of the program itself, but to severe damages of the entire robot.

To mitigate this shortcoming, most commercial manufacturers of industrial robots provide domain-specific languages for their robot systems. These languages are usually text-based and offer the possibility to declare robotic-specific data types, to specify simple motions, and to interact with tools and sensors via I/O operations. Examples are the KUKA Robot Language or RAPID from ABB. To run such a program, it is transmitted to the robot controller, where it is executed obeying real-time constraints. These languages focus on a narrow set of required features and prohibit potentially dangerous operations such as direct memory access.
However, a major drawback of these languages is the lack of extensibility. They are tailored to the underlying controller and, as a consequence, only offer a fixed, controller-specific set of instructions. Due to this design, even robot manufacturers have difficulties extending these languages with new instructions or adapting them to novel, challenging requirements such as the synchronization of cooperating robots or the tight integration of sensor-guided motions.

For that reason, we have developed a novel software architecture~\cite{Hoffmann2009} for programming industrial robots in the research project \emph{SoftRobot}. Its core part is the \emph{Robotics API} which represents an extensible object-oriented framework for robotic applications (available in Java and C\#). 
Similar to robotic DSLs, the Robotics API offers an abstraction from real-time programming facilitating the development of robotic software. With a general-purpose programming language and a robotic-specific application programming interface (API), rapid development of domain-specific languages for industrial robots
is possible. In contrast to existing robotics programming languages, we intend to develop lightweight DSLs, that is, languages which are implemented using well-defined interfaces and without other dependencies on the underlying robot system. As a proof-of-concept, we decided to reverse-engineer the KUKA Robot Language (KRL) and implement it on top of our architecture. We have chosen KRL to show that we can achieve a similar expressiveness as \emph{classical} robot programming languages and simulate their execution semantics. Moreover, we can execute legacy code written in KRL this way and thus maintain a certain degree of compatibility.

The paper is structured as follows: In Sect.~\ref{sec:krl}, the KUKA Robot Language is introduced and its main features are explained using an example. Subsequently, Sect.~\ref{sec:rapi} describes the Robotics API and our software architecture which was used to implement a lightweight interpreter for KRL. The steps required for reverse-engineering KRL are explained in Sect.~\ref{sec:approach}. While static preprocessing of KRL programs is described in Sect.~\ref{sec:preprocessing}, different approaches for interpreting KRL on top of our software architecture are discussed in Sect.~\ref{sec:interpretation}. Finally, Sect.~\ref{sec:conclusion} concludes the paper.

\section{KUKA Robot Language} \label{sec:krl}

Today, every KUKA industrial robot is programmed using the proprietary \emph{KUKA Robot Language} which is an imperative programming language similar to \emph{Pascal}. In addition to typical programming statements such as variable assignments, conditionals and loops, KRL provides robotic-specific statements e.g. for specifying motions and interacting with tools. Moreover, KRLs execution semantics uses two program counters to implement a feature known as \emph{advance run}. The second counter is usually ahead of the regular program counter and is used for planning motions in advance. In particular, this is used to blend two subsequent motions. An extensive reference to KRL can be found in~\cite{KUKA:SystemSoftware55}.

An example program showing the main features of KRL is given in Fig.~\ref{fig:krl}. Like all KRL programs, it contains a declaration section followed by a statements section which starts with the first statement and comprises the actual program. Every variable must be declared in the declaration section and can later be assigned in the statements sections. Besides common data types (\krl{INT}, \krl{REAL}, \krl{BOOL} and \krl{CHAR}), KRL allows the definition of arrays and user-defined structs and provides predefined structs such as \krl{AXIS} and \krl{POS} which are specific to robotics. There is a large number of predefined global variables, e.g. for setting the max. velocity of the robot's joints (\krl{$VEL_AXIS}) or for accessing binary inputs (\krl{$IN}) and outputs (\krl{$OUT}). To control program flow, KRL has several statements such as conditionals (e.g. \krl{IF}, \krl{SWITCH}), loops (e.g. \krl{FOR}, \krl{REPEAT ... UNTIL}, \krl{WHILE}) and jumps (\krl{GOTO}).

\begin{figure}%
\begin{lstlisting}
DEF example()
	DECL INT i
	DECL POS cpos
	DECL AXIS jpos
	
	FOR i=1 TO 6
		$VEL_AXIS[i]=60
	ENDFOR
	
	jpos = {AXIS: A1 0,A2 -90,A3 90,A4 0,A5 0,A6 0}
	PTP jpos

	IF $IN[1] == TRUE THEN
		cpos = {POS: X 300,Y -100,Z 1500,A 0,B 90,C 0}
	ELSE
		cpos = {POS: X 250,Y -200,Z 1300,A 0,B 90,C 0}
	ENDIF
	
	INTERRUPT DECL 3 WHEN $IN[2]==FALSE DO BRAKE
	INTERRUPT ON 3
	
	TRIGGER WHEN DISTANCE=0 DELAY=20 DO $OUT[2]=TRUE
	LIN cpos
	LIN	{POS: X 250,Y -100,Z 1400, A 0,B 90,C 0} C_DIS
	PTP jpos
	
	INTERRUPT OFF 3
END
\end{lstlisting}
\caption{Example program written in KRL.}
\label{fig:krl}%
\end{figure}

\begin{figure}[htbp]
	\centering
		\includegraphics[height=0.6in]{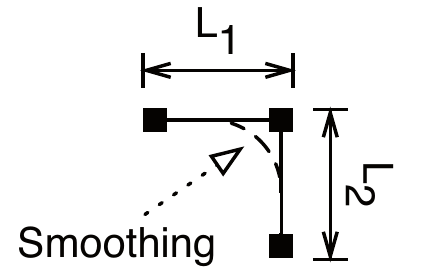}
	\caption{Two blended linear motions, $L_1$ and $L_2$, with n automatically generated transition.}
	\label{fig:blending}
\end{figure}

Because KRL is a robotic DSL, it has built-in statements for programming motions such as \krl{PTP} or \krl{LIN}. While \krl{PTP} executes a point-to-point motion to its programmed end point, the \krl{LIN} statement executes a linear motion to a specified target in Cartesian space. Besides these two motion types, KRL currently supports circular motions and splines. Using the \emph{advance run} feature, motions are pre-calculated and planned before the actual program execution reaches the corresponding motion statement. This allows blended movements, where the motion does not stop exactly at the programmed target but blends into the subsequent motion (cf.~Fig.~\ref{fig:blending}). Usually, this approach leads to faster execution of multiple motion statements by reducing acceleration and deceleration phases. To indicate that a target position does not need to be reached exactly, an additional parameter (e.g. \krl{C_DIS} in the example program) is used in the corresponding motion statement. However, there are KRL statements like reading an input which stop the advance run making it impossible to blend between two motions~\cite{KUKA:SystemSoftware55}.

In KRL, the communication with external tools and sensors is usually done by using digital or analog in- and outputs. Hence, KRL provides predefined global variables (cf. variables \krl{$IN} and \krl{$OUT}) and additional statements to handle in- and outputs. These variables and statements can be used in a \krl{TRIGGER} statement to define path-related switching actions. In the example from Fig.~\ref{fig:krl}, the binary output \krl{$OUT[2]} is set to true 20ms after the linear motion to \krl{cpos} was started. Another feature of KRL are interrupts, where the current program is suspended  when an event (e.g. an input)  occurs and a subprogram is executed. The event and the subprogram are defined by an \krl{INTERRUPT} declaration statement. After being defined, the interrupted can be switched on and off. In the example program above, an interrupt is declared, which fires when the binary input \krl{$IN[2]} becomes true. The interrupt-subprogram causes the robot to brake immediately (\krl{BRAKE}). Besides the \emph{submit interpreter}~\cite{KUKA:SystemSoftware55}, which allows running a special second program in parallel with the main program, triggers and interrupts are the only means to express concurrency in KRL -- albeit in a limited manner.


\section{Robotics API} \label{sec:rapi}

\begin{figure}
	\centering
		\includegraphics[width=0.75\columnwidth]{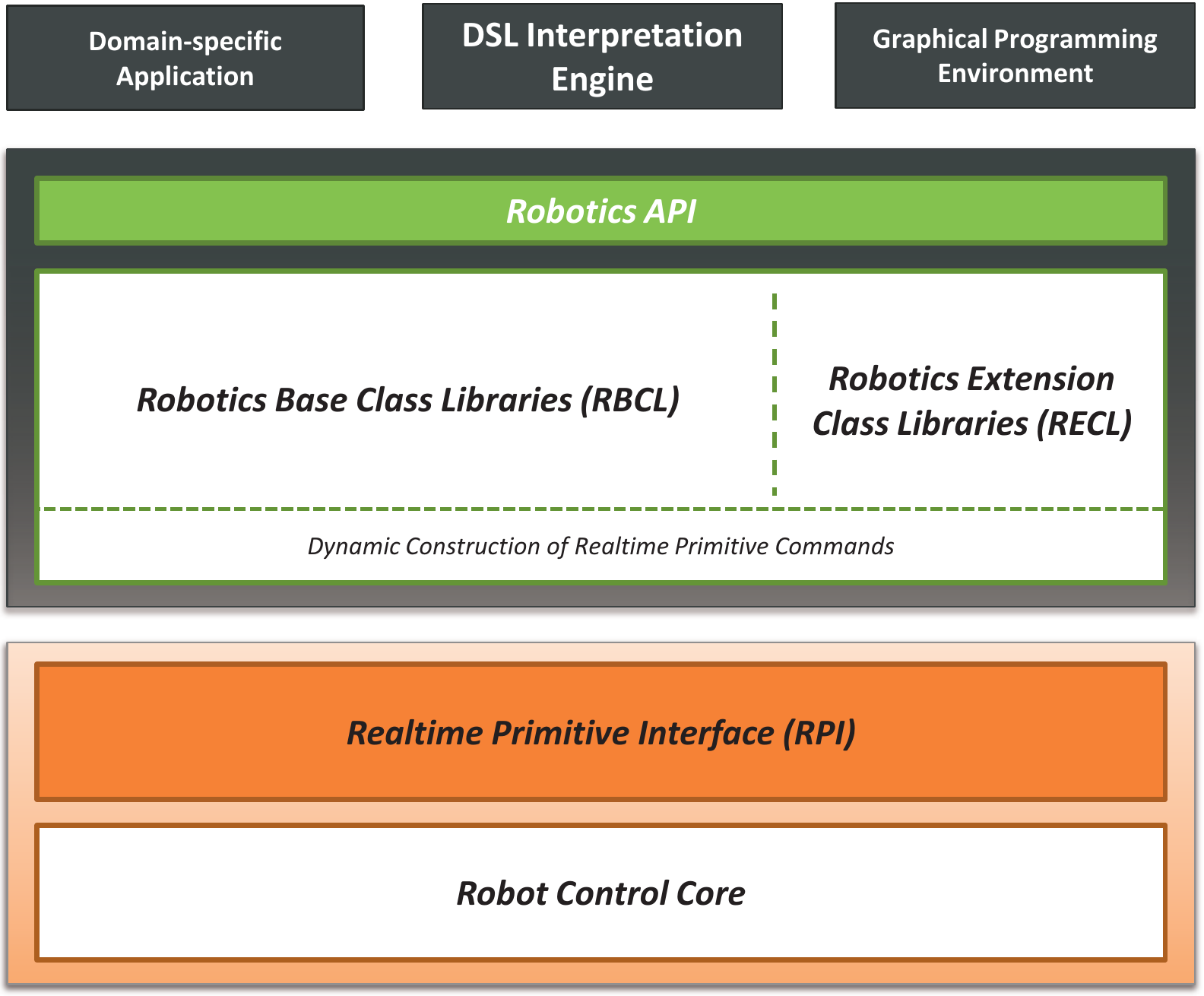}
	\caption{The Robotics API and its context in the software architecture.}
	\label{fig:architecture}
\end{figure}

Fig.~\ref{fig:architecture} illustrates the important parts of the software architecture introduced in~\cite{Hoffmann2009}. The central concept is that real-time critical commands can be specified using the object-oriented \emph{Robotics API} and are then translated into commands for the \emph{Realtime Primitives Interface}~\cite{Vistein2010}. The \emph{Robot Control Core (RCC)} executes these commands, which consist of a combination of pre-defined (yet extendable) calculation modules and a specification of the data flow among them. Implementing only the RCC with real-time aspects in mind is sufficient to allow the atomic execution of Robotics API commands while obeying real-time constraints. Here, only the aspects of the programming model relevant to this work will be introduced. A comprehensive overview of the programming interface can be found in~\cite{Angerer2010}.

The Robotics API is implemented as a library on top of a general purpose programming language. 
In terms of concepts, the Robotics API covers the same scope that the KUKA Robot Language does (see Sec.~\ref{sec:krl}), offering robotic specific data types, movement and tool commands as well as control flow structures as provided by the host language. Additionally, the Robotics API incorporates a comprehensive robotics domain model that extends beyond the set of specific data types that KRL provides. This simplifies the development of diverse, complex robotic applications, which is also supported by reusable class structures and reusable logic.
However, runtime environments of common object-oriented programming languages do not guarantee real-time constraints during program execution. Thus, reacting to certain sensor events cannot be done in real-time using eventing concepts built into the programming language.

\begin{figure}[htbp]
	\centering
		\includegraphics[width=0.85\columnwidth]{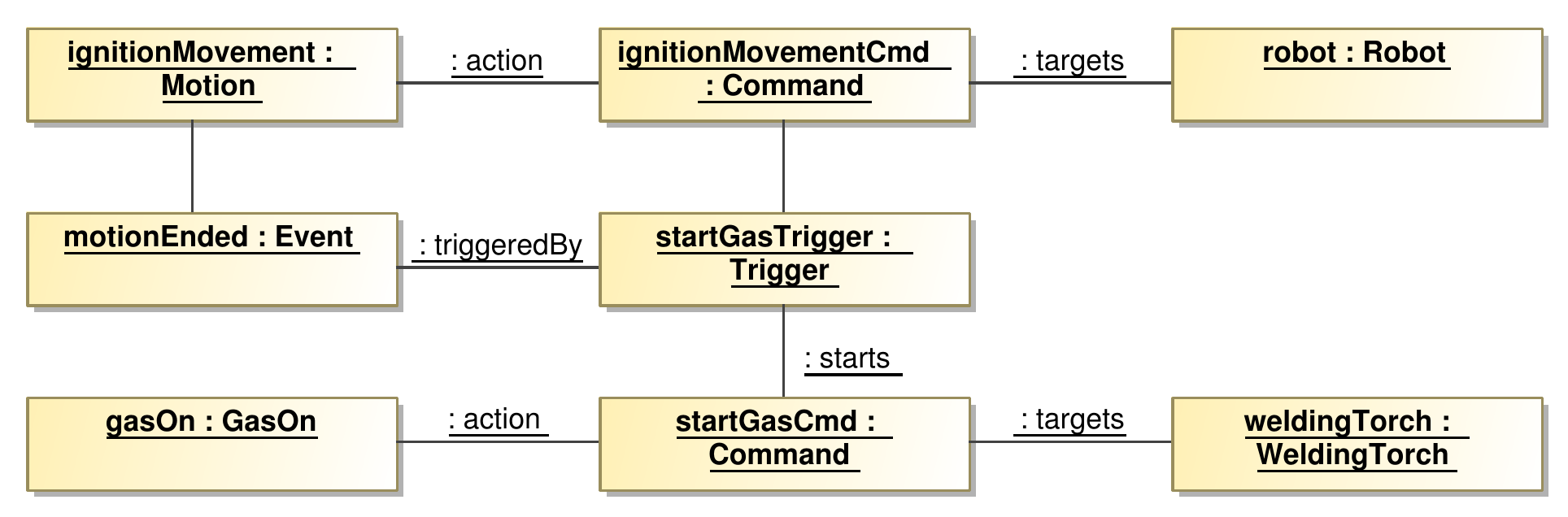}
	\caption{Robotics API command}
	\label{fig:beforemapping}
\end{figure}

To meet real-time requirements, e.g. for motion blending or event handling, commands in the Robotics API can be linked with a special mechanism prior to execution. Commands that are combined in that way are executed atomically on the  Robot Control Core.  Fig.~\ref{fig:beforemapping} shows an example of such a combined command structure. This example is taken from a welding application and specifies the initialization of the welding gas flow as soon as a robot motion has ended. To achieve this, the motion (titled \emph{ignitionMovement}) defines a special event (\emph{motionEnded}) that is used as a trigger (\emph{startGasTrigger}) for starting the subsequent command (\emph{startGasCommand}). Such a command structure is converted into a dataflow specification that is executable by the RCC with real-time guarantees. This, however, is a different concept of handling events occurring during execution than the concept employed by KRL, where it is possible to call arbitrary methods when an event occurs.

Additionally, the way commands for robots or tools are processed inside the Robotics API differs from how they are handled in the KUKA Robot language. KRL interprets program instructions ahead of their actual execution which allows for successive blending between subsequent motions (see Sec.~\ref{sec:krl}). In today's general purpose programming languages, which the Robotics API is intended to be used with, such an ahead-of-time interpretation is usually not provided. 

Our transaction-like mechanism for specifying real-time critical commands is suitable for covering the requirements in almost any domain of industrial robotics today~\cite{Hoffmann2009}. It also allows for blending between subsequent motions, but only for a finite set of motions and not incrementally. However, the execution semantics clearly differs from those of the KUKA Robot Language, which constitutes an additional challenge for interpreting legacy KRL programs on top of the Robotics~API.


\section{Approach} \label{sec:approach}

The goal of this work was running KRL in an entirely different execution environment and with a different underlying execution paradigm. The first problem is that KRL is only documented in user manuals, i.e. there is no publicly available formal specification or grammar of the language. Semantics is only documented by informal descriptions. Additionally, even if it is possible to sufficiently determine the semantic meaning of every construct in the language, a successful reimplementation must not only retain the same control flow as the original implementation but also ensure that robot movements are planned and executed the same way. This task is difficult, because of different underlying concepts of KRL and the Robotics API used to implement motion commands. KRL can start a motion and continue the execution of the program while the motion is performed by the robot. Commands read while continuing the execution can influence the motion (cf. \emph{advance run} in Sect.~\ref{sec:krl}). Replicating this in a simple way requires a means of augmenting a motion which has already been started, which is conceptually incompatible with the Robotics~API.

Starting with a simple benchmark program written in KRL and the programming manual~\cite{KUKA:SystemSoftware55}, every command found in either code or the handbook was added to a list of valid KRL statements. To find the set of allowed parameters for each command, the manual proved to be insufficient, even though it uses EBNF-like notation to explain some commands. The set of parameters specified in the manual is often either incomplete or vague.
To counter the problems stemming from a lack of documentation, we extracted about one megabyte of valid KRL source code from a KUKA robot control unit. This in itself is usually not enough to construct a grammar for a programming language in a reasonable amount of time. Combined with a very restrictive preliminary grammar built manually using the handbook mentioned earlier, the source code proved to be a valuable asset. The preliminary grammar was constructed in a conservative manner: Whenever the syntax was not entirely clear in the handbook, the most restrictive assumption was made to construct the corresponding grammar rule. After having finished the construction, both lexer and parser were generated and extended to simplify finding missing or insufficient grammar rules. All common constructs not included in the grammar so far were added. If a specific parse error happened only with a single source file, the official KRL implementation KUKA.OfficeLite~\cite{KUKA:OfficeLite} was used to validate the construct before it was added to the grammar. This way a comprehensive grammar for KRL was built from scratch in a short amount of time.

Having obtained a usable grammar, we next implemented a way of \emph{mapping} KRL to the Robotics API introduced in Sect.~\ref{sec:rapi}. Although translating KRL to Java with calls to the Robotics API is possible, we chose to build an interpreter for the DSL. The official KUKA implementations of KRL offer some debugging options and features like backward execution of movement commands. Trying to mimic these features in a translation-based approach would complicate the translation process. An interpreter, on the other hand, allows a high level of interaction with the interpreted program. It simulates a CPU but can be accessed and manipulated easier than the physical CPU. Further advantages of the interpretation approach will be detailed in Sect. \ref{sec:interpretation}.

\section{Preprocessing} \label{sec:preprocessing} 
From the grammar, both lexer and parser were generated using ANTLR~\cite{Parr1995} (\emph{ANother Tool for Language Recognition}). We enriched the grammar with the description of a tree structure that the generated parser uses to output an Abstract Syntax Tree (AST) instead of a parse tree. A parse tree is a representation of which rules were used to parse the input whereas an AST is less verbose and represents only the information necessary for the purpose of interpretation.

Since KRL is both statically and explicitly typed, comprehensive type-checking is possible before actually executing a program. This is advantageous, since having to terminate a program because of an invalid assignment while in the middle of the execution is discouraged, especially in a DSL that is used to control heavy machinery. Type checking requires all symbols and their types to be known. Additionally, it is paramount that symbol scoping, that is, different visibility in different areas of the program, is established. Thus, a symbol table has to be generated before a type checker can be run. 

\begin{figure}[tbp]
	\centering
		\includegraphics[width=0.9\columnwidth]{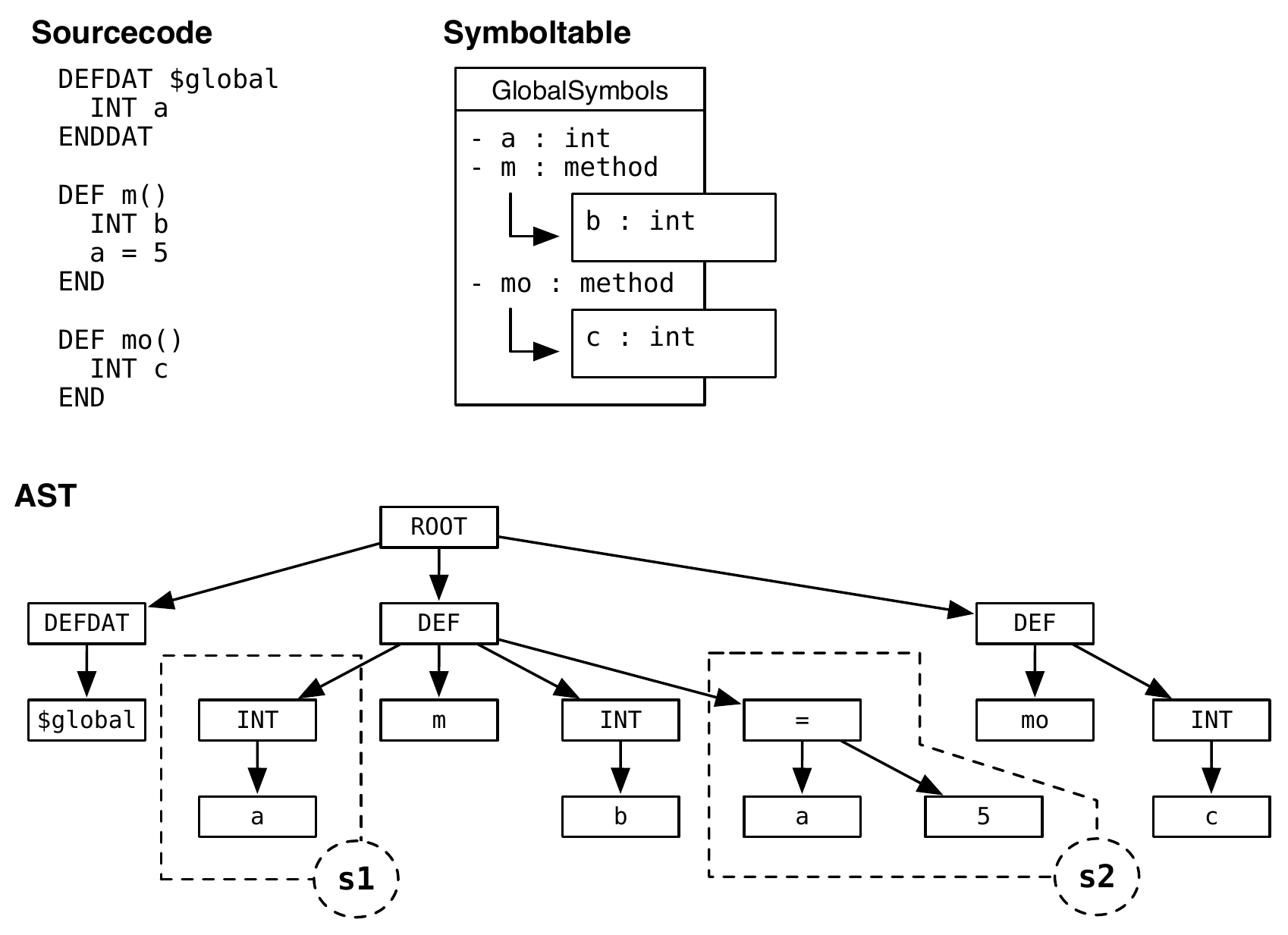}
	\caption{KRL sourcecode, AST generated by the parser and associated symbol table hierarchy.}
	\label{fig:ast}
\end{figure}

The program that fills the symbol table works on the AST created by the parser -- an example is displayed in Fig. \ref{fig:ast}. A hierarchy of symbol tables is established to allow for easy integration of scoping: only symbols are visible that are in a symbol table which is an ancestor to the current symbol table. In this context, \emph{current} symbol table means local to the expression for which visibility of a symbol has to be determined. If a statement in a method $m$ tries to read a variable $a$, the symbol table used to collect symbols of said method is the \emph{current} symbol table. In the symbol table hierarchy given at the top of Fig. \ref{fig:ast}, variable $a$ is thus visible to method $m$, since it is in a symbol table that is an ancestor of the current symbol table for statements in method $m$. 

After having collected all symbols in a given KRL program, type checking is trivial: The AST has to be read a second time and every expression has to be checked for type compatibility. Type checking can be done by verifying that a set of type constraints is met \cite{dragonbook}. Considering the source code in Fig. \ref{fig:ast}, we must verify that the type of the variable on the left side of the assignment is compatible with the type of the expression on the right. In the context of an assignment, two types are considered compatible if lossless conversion between the type of the expression and the variable is possible. A set of similar constraints was found for all expressions in KRL and implemented in the type checker. The type checker also adds type information to the AST to avoid having to compute this information again in the interpreter.

\section{Interpretation} \label{sec:interpretation}

Traversing the AST is a simple way of executing a DSL. We will call this approach \emph{tree-based interpretation}. It is a straightforward implementation of an interpreter, but not the only feasible method. Literature suggests that a different approach, called \emph{bytecode interpretation}, is also widely used \cite{LangImplPatterns}. Since our implementation using a tree-based approach is both more mature and versatile than our implementation of a bytecode interpreter, we will focus tree-based interpretation first. Later in this chapter, a bytecode-based prototype will be introduced and compared to the original approach.

\begin{figure}[tbp]
	\centering
		\includegraphics[width=0.78\columnwidth]{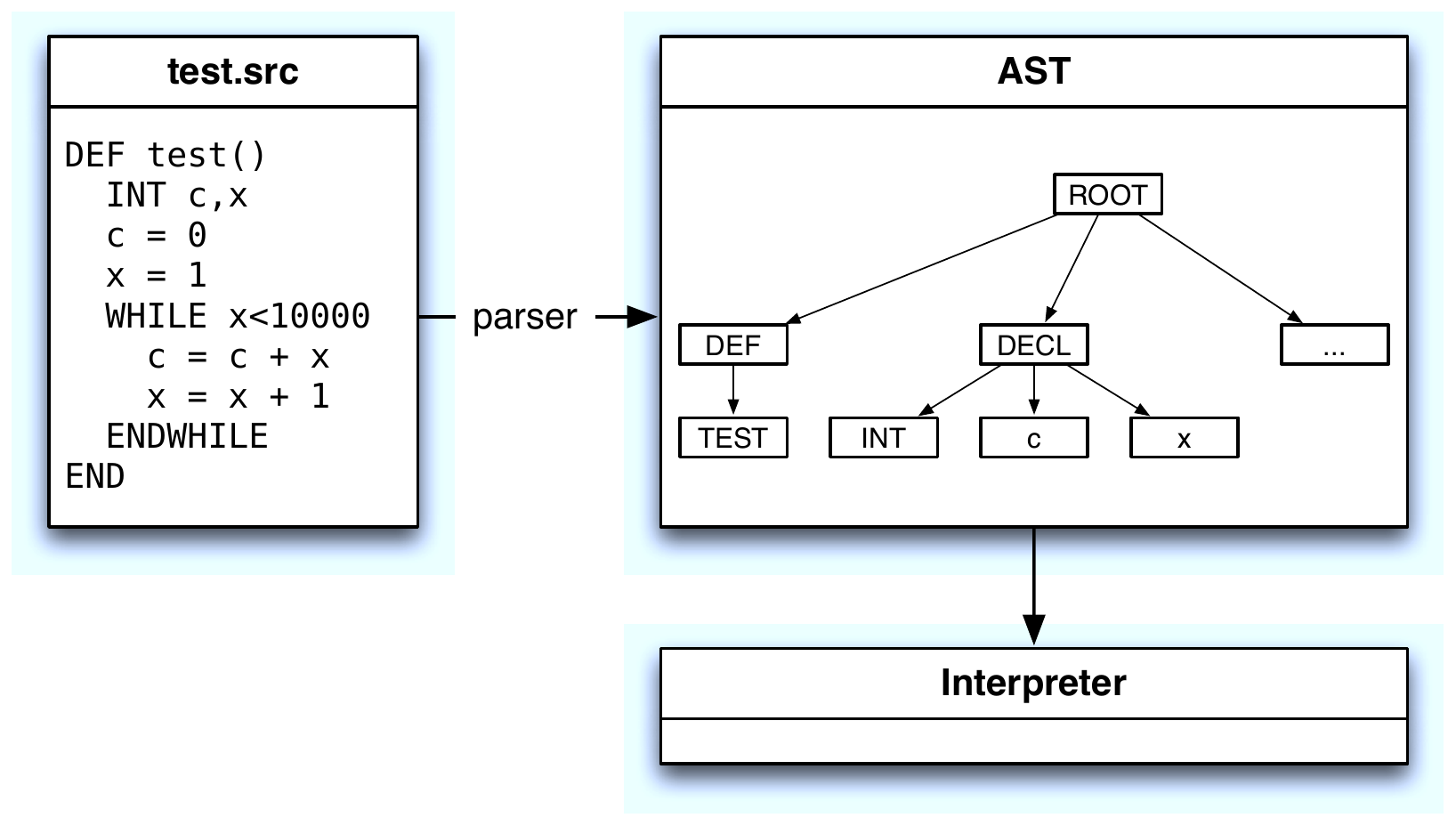}
	\caption{Architecture when using a tree-based interpreter.}
	\label{fig:tree-architecture}
\end{figure}

Tree-based interpretation relies on the idea that a KRL program can be effectively executed by walking the AST generated from it and executing each node in it as shown in Fig. \ref{fig:tree-architecture}. For simple statements, like a variable declaration (see Fig.~\ref{fig:ast}, label $s1$), execution comes down to reading the declaration node, reading the variable name in the child node and allocating a corresponding amount of memory. For more complex statements like an assignment (see Fig.~\ref{fig:ast}, label $s2$), execution requires executing both child-nodes. The value returned by executing the second child-node -- and thus the right side of the assignment -- needs to be assigned to the position in memory referenced by the first child-node.

Control flow constructs like loops are simple to execute: for an unconditional loop, its body is executed iteratively until a statement which causes the loop to exit is encountered. However, a serious problem arises with executing statements that severely impact control flow (e.g. a return statement, a procedure, a function call, or a \krl{GOTO}). It is sufficient to focus on the most difficult of these statements, \krl{GOTO}, since the others statements can be simulated with it. In a tree-based interpreter, two stacks need to be maintained: The call stack of the interpreter (interpreter call stack) itself and the call stack of the program being interpreted (program call stack). When interpreting a \krl{LOOP} for instance, the interpreter call stack changes because a method is invoked that handles the execution of the loop. The program call stack remains the same as it only changes with method calls and jumps in KRL cannot target a label outside the current method. 

Consider a \krl{GOTO} statement inside a loop which jumps to a statement outside said loop. A jump like that would require the method executing the \krl{LOOP} statement to return execution to its caller and tell it to continue at the target label of the jump. This requires reversing the call stack, and transferring information on where to continue execution to the caller -- in complex cases through a call-hierarchy of arbitrary height. Although this can be achieved by wrapping every method used for node execution, a more elegant albeit unusual approach is using the exception concept provided by most high-level programming languages, as suggested in \cite{LangImplPatterns}. Exceptions can pass through the call hierarchy and be caught in appropriate places with a minimum of additional code. Additionally, since none of the debug information carried by the exception is required, cost for exception creation is minimal, as suggested by \cite[p. 174]{572520}.

At the beginning of this chapter, a different approach, bytecode interpretation, was mentioned. Compared to a tree-based interpreter, a bytecode interpreter does not walk a complex data structure like a tree but instead works on simple code consisting of small \emph{atomic} operations. Using a less verbose data structure leads to less overhead in the interpreter. Extracting an operation and its parameters from the AST requires accessing different nodes of the tree and thus accessing different addresses in memory using pointers. Bytecode can be constructed in a way that groups operations and parameters together, making the cost of accessing an operation and its parameters as cheap as reading a byte-sequence from memory.

\begin{figure}[tbp]
	\centering
		\includegraphics[width=0.78\columnwidth]{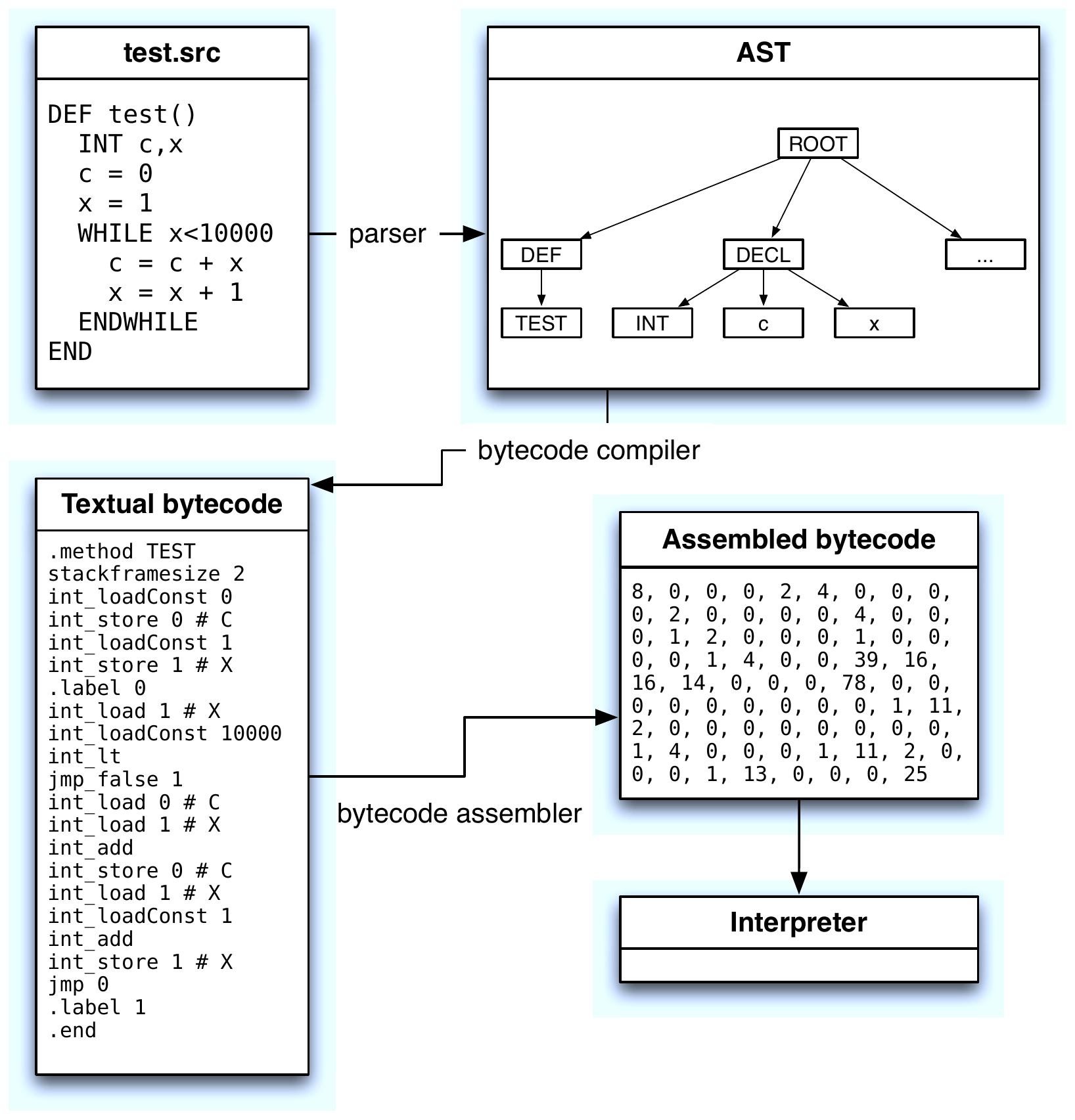}
	\caption{Process used to generate bytecode from a KRL program.}
	\label{fig:bytecode-architecture}
\end{figure}

The bytecode has to be generated out of the KRL program before interpretation can start. This is achieved by reusing the annotated AST introduced in Sect.~\ref{sec:approach}. Since bytecode is usually represented as an array of byte values, a standard approach is to first generate human readable bytecode that is easy to read and debug, and then assemble the textual representation into an actual byte-array. We designed each bytecode operation to be as simple as possible. Required memory is determined by the compiler and numerical indexes are used to address variables. This way, memory allocation is dealt with before the program is executed, resulting in a leaner interpreter. The translation process and the results generated in each step are displayed in Fig. \ref{fig:bytecode-architecture}.


The Robotics API is used to execute movement commands in both the tree-based interpretation approach as well as the bytecode interpretation approach. As explained in Sect.~\ref{sec:krl}, KRL can blend motions together, thus making the transition between two separate motions smooth, as illustrated in Fig.~\ref{fig:blending}. To achieve this, KRL executes motions asynchronously and continues interpretation until it finds the next motion to be executed. When found, a soft transition between the running and the next motion is calculated and applied. Mapping this behavior to the Robotics API is non-trivial, because the way a motion is executed cannot be influenced once execution has started. To mimic KRL, we do not start execution of a motion as soon as it is found. Instead interpretation is continued and all motion commands are queued until a motion is encountered that may not be blended. This can either be the case when a motion has a parameter indicating that it is not supposed to blend with a subsequent motion, when a special command prevents blending as mentioned in Sect.~\ref{sec:krl} or when the queue is full. If any of the above applies, the motion-queue is executed with blending between each motion. The rationale behind this is that interpretation of control flow and computational statements takes only a minor amount of time. Thus, the delay introduced by deferred execution of motion commands is minimal, but allows us to closely mimic the execution semantics of KRL.

Interrupts in KRL halt execution of the main program and run a subroutine when an event condition becomes true. An interpreter can mimic this behavior by a separate thread that periodically checks all event conditions. When a condition becomes true, the main thread executing the KRL program has to block and a new thread executing the interrupt subroutine has to be created. Once the subroutine has finished, the main thread can resume. Apart from this scenario, more complex interrupt invocations are possible: an interrupt can fire while an interrupt subroutine is being executed. To handle both scenarios, all threads executing either the main program or an interrupt subroutine are managed on a stack. Only the thread on top of the stack is allowed to execute statements, all other threads need to block. When an interrupt subroutine finishes, the thread executing it is removed from the stack, allowing another interrupt to finish. This concept can cope with an arbitrary number of interlaced interrupts without explicitly suspending threads and makes resource synchronization unnecessary. Additionally, the same mechanism can be used to implement triggers: instead of monitoring a condition, the robots position is considered. When the robot reaches the point specified in the trigger, its subroutine is executed just like an interrupt subroutine. However, this approach does not guarantee any real-time constraints for both interrupts and triggers. Alternatively, simple triggers (e.g. setting outputs) that meet real-time requirements can be established by the transaction-like mechanism of the Robotics API (cf. Sect.~\ref{sec:rapi}).

The third way of adding concurrency is the submit interpreter, which allows executing a different program in parallel to the main KRL program. Implementing the submit interpreter is possible by running two separate interpreter instances concurrently on the same memory. Because KRL documentation does not specify how the submit interpreter is scheduled and whether it receives CPU time, we have decided not to include an implementation in our prototypes.

\section{Conclusion} \label{sec:conclusion}


In this paper, we have shown that it is possible to run KRL using the Java-based Robotics API developed in the \emph{SoftRobot} project. Running KRL is necessary in order to support legacy applications and additionally serves as proof that our approach is indeed viable. We have introduced the Robotics API and shown that it is superior to KRL in many areas, but also that its semantics differs in other areas. Additionally. we have shown that reverse-engineering KRL is possible and that a grammar for the language can be build using only information from the KRL manual and results from trying statements in KUKA.OfficeLite. 

To execute KRL, we have introduced two separate approaches, each with a different set of advantages and thus use cases. 
Comparing both approaches, the tree-based interpreter is flexible because it works on a more verbose data-structure. This makes adding features like debugging simple, because debug information is directly available to the interpreter. The bytecode interpreter works on a lean data-structure including only information vital to executing the program. Furthermore, bytecode can be represented as an array making dereferencing pointers and thus random memory access unnecessary. Because of this, more instructions fit into the CPU cache and execution is faster. We tested performance while executing a simple benchmark written in KRL. With both interpreters written in Java, the bytecode interpreter is about six times faster than the tree-based interpreter on all major platforms (Microsoft Windows Vista, Ubuntu Linux, Mac OS X) tested on the same machine. Furthermore, a bytecode interpreter prototype written in C is up to 300 times faster than the tree-based interpreter written in Java. However, interfacing the bytecode interpreter written in C with the Java-based API is not possible yet.

The Robotics API was interfaced with the interpreters to implement robot commands like motions and to gain access to I/Os. We assumed that executing control flow and calculation statements takes only a small amount of time compared to the time it takes the robot to complete a motion command. Under that assumption, we were able to show that differences in semantics between the API and KRL can be bridged by batch-executing motions. Concurrency as offered by KRL can be added to our interpreters by using threads and a stack-based synchronization mechanism.

Future research will focus on extending and modifying KRL to include features like synchronization between robots. The interpreters introduced in this paper were designed to be extendable and can be used for prototyping additions to the original DSL. Furthermore, more specific commands, e.g. for welding applications, will be integrated. So far, the Robotics API does not include a way of specifying complex triggers like it is possible in KRL. Embedding a simple DSL to specify triggers in and using a lean bytecode interpreter to execute those triggers could be a way of mitigating this shortcoming and will be a topic of future research.


\section*{Acknowledgement}

We would like to thank Prof. Terrence Parr for his important remarks on interpreting KRL.

\bibliographystyle{IEEEtran}
\bibliography{custom}

\begin{thebibliography}{10}
\providecommand{\url}[1]{#1}
\csname url@samestyle\endcsname
\providecommand{\newblock}{\relax}
\providecommand{\bibinfo}[2]{#2}
\providecommand{\BIBentrySTDinterwordspacing}{\spaceskip=0pt\relax}
\providecommand{\BIBentryALTinterwordstretchfactor}{4}
\providecommand{\BIBentryALTinterwordspacing}{\spaceskip=\fontdimen2\font plus
\BIBentryALTinterwordstretchfactor\fontdimen3\font minus
  \fontdimen4\font\relax}
\providecommand{\BIBforeignlanguage}[2]{{%
\expandafter\ifx\csname l@#1\endcsname\relax
\typeout{** WARNING: IEEEtran.bst: No hyphenation pattern has been}%
\typeout{** loaded for the language `#1'. Using the pattern for}%
\typeout{** the default language instead.}%
\else
\language=\csname l@#1\endcsname
\fi
#2}}
\providecommand{\BIBdecl}{\relax}
\BIBdecl

\bibitem{Biggs2003}
G.~Biggs and B.~MacDonald, ``A survey of robot programming systems,'' in
  \emph{Proc. Australasian Conference on Robotics and Automation}, J.~Roberts
  and G.~Wyeth, Eds., Brisbane, Australia, Dec. 2003.

\bibitem{Hoffmann2009}
A.~Hoffmann, A.~Angerer, F.~Ortmeier, M.~Vistein, and W.~Reif, ``Hiding
  real-time: A new approach for the software development of industrial
  robots,'' in \emph{Proc. 2009 IEEE/RSJ Intl. Conf. on Intelligent Robots and
  Systems}, St. Louis, MO, USA, Oct. 2009, pp. 2108--2113.

\bibitem{KUKA:SystemSoftware55}
\emph{KUKA System Software 5.5 -- Operating and Programming Instructions for
  System Integrators}, {KUKA} {R}oboter {G}mbH, Augsburg, Germany, 2009.

\bibitem{Vistein2010}
M.~Vistein, A.~Angerer, A.~Hoffmann, A.~Schierl, and W.~Reif, ``Interfacing
  industrial robots using realtime primitives,'' in \emph{Proc. 2010 IEEE Intl.
  Conf. on Automation and Logistics}, Hong Kong, China, Aug. 2010.

\bibitem{Angerer2010}
A.~Angerer, A.~Hoffmann, A.~Schierl, M.~Vistein, and W.~Reif, ``The {Robotics
  API}: {A}n object-oriented framework for modeling industrial robotics
  applications,'' in \emph{Proc. 2010 IEEE/RSJ Intl. Conf. on Intelligent
  Robots and Systems}, Taipei, Taiwan, Oct. 2010.

\bibitem{KUKA:OfficeLite}
\BIBentryALTinterwordspacing
{KUKA}.{O}ffice{L}ite. {KUKA} {R}oboter {G}mbH. [Online]. Available:
  \url{http://www.kuka-robotics.com/en/products/software/kuka_sim/kuka_sim_det%
ail/}
\BIBentrySTDinterwordspacing

\bibitem{Parr1995}
T.~Parr and R.~Quong, ``{ANTLR: A predicated-LL(k) parser generator},''
  \emph{Software-Practice and Experience}, vol.~25, no.~7, pp. 789--810, 1995.

\bibitem{dragonbook}
A.~V. Aho, M.~S. Lam, R.~Sethi, and J.~D. Ullman, \emph{Compilers: Principles,
  Techniques, and Tools}, 2nd~ed.\hskip 1em plus 0.5em minus 0.4em\relax
  {Addison Wesley}, 2006.

\bibitem{LangImplPatterns}
T.~Parr, \emph{Language Implementation Patterns: Create Your Own
  Domain-Specific and General Programming Languages}.\hskip 1em plus 0.5em
  minus 0.4em\relax Pragmatic Bookshelf, 2009.

\bibitem{572520}
J.~Shirazi, \emph{Java Performance Tuning}.\hskip 1em plus 0.5em minus
  0.4em\relax Sebastopol, CA, USA: O'Reilly \& Associates, Inc., 2002.

\end{thebibliography}

\end{document}